\documentclass[runningheads]{cl2emult}
\usepackage{makeidx}  
\usepackage{graphicx} 
\usepackage{math}     
\makeindex            

\begin{document}
\title*{A Flexible Rule Compiler for Speech Synthesis}
\toctitle{A Flexible Rule Compiler for Speech Synthesis}
%
%
\titlerunning{A Flexible Rule Compiler for Speech Synthesis}
%
\author{Wojciech Skut
\and Stefan Ulrich
\and Kathrine Hammervold}
\authorrunning{Skut et al.}
%
%
\institute{
     Rhetorical Systems, 4 Crichton's Close, Edinburgh, EH8 8DT, Scotland
}

\bibliographystyle{apalike}

\maketitle              

\begin{abstract}
We present\index{abstract} a flexible rule compiler developed for a
text-to-speech (TTS) system. The compiler converts a set of rules into
a finite-state transducer (FST). The input and output of the FST are
subject to parameterization, so that the system can be applied to
strings and sequences of feature-structures. The resulting transducer
is guaranteed to realize a function (as opposed to a relation), and
therefore can be implemented as a deterministic device (either a
deterministic FST or a bimachine).
\end{abstract}

\section{Motivation}

Implementations of TTS systems are often based on operations
transforming one sequence of symbols or objects into another. Starting
from the input string, the system creates a sequence of tokens which
are subject to part-of-speech tagging, homograph disambiguation rules,
lexical lookup and grapheme-to-phoneme conversion.  The resulting
phonetic transcriptions are also transformed by syllabification rules,
post-lexical reductions, etc.

The character of the above transformations suggests finite-state
transducers (FSTs) as a modelling framework
\cite{Sproat:96,Mohri:97}. However, this is not always straightforward
for two reasons.

Firstly, the transformations are more often expressed by rules than
encoded directly in finite-state networks. In order to overcome this
difficulty, we need an adequate compiler converting the rules into an
FST.

Secondly, finite-state machines require a finite alphabet of symbols
while it is often more adequate to encode linguistic information using
structured representations (e.g. feature structures) the inventory of
which might be potentially infinite. Thus, the compilation method must
be able to reduce the inifinite set of feature structures to a finite
FST input alphabet.

In this paper, we show how these two problems have been solved in
rVoice, a speech synthesis system developed at Rhetorical Systems.

\section{Definitions and Notation}

A {\em deterministic finite-state automaton (acceptor, DFSA)}  over
a finite alphabet $\Sigma$ is a quintuple
$A=(\Sigma, Q, q_0,\delta, F)$ such that:

\begin{description}
\item[$Q$] is a finite set of states, and $q_0 \in Q$ is the initial state of $A$;
\item[$\delta: Q \times \Sigma \rightarrow Q$] is the transition function
of $A$; 
\item[$F \subset Q$] is a non-empty set of final states.
\end{description}
A {\em (non-deterministic) finite-state transducer} (FST) over an
input alphabet $\Sigma$ and an output alphabet $\Delta$ is a 6-tuple
$T=(\Sigma, \Delta, Q, I, E, F)$ such that:

\begin{description}
\item[$Q$] is a finite set of states;
\item[$I \subset Q$] is the set of initial  and $F \subset Q$ that of final states;
\item[$E\subset Q\times Q \times \Sigma\cup\{\epsilon\}\times
\Delta^*$] is the set of transitions of $T$. We call a quadruple
$(q,q',a,o)\in E$ a transition from $q$ to $q'$ with input $a$ and
output $o$.
\end{description}
 Each transducer $T$ defines a relation $R_T$ on $\Sigma^* \times
\Delta^*$ such that $(s,o) \in R_T$ iff there exists a decomposition
of $s$ and $o$ into substrings $s_1,\ldots, s_t$, $o_1,\ldots,o_t$
such that $s_1\cdot\ldots\cdot s_t=s$, $o_1\cdot\ldots\cdot o_t=o$ and
there exist states $q_0\ldots q_t \in Q$, $q_0 \in I, q_t\in F$, such
that $(q_{i-1}, q_i, s_i, o_i)\in E$ for $i =1\ldots t$.

If $R_T$ is a (partial) function from $\Sigma^*$ to $\Delta^*$, 
the FST is called {\em functional}.

A {\em deterministic finite-state transducer} (DFST) is a DFSA whose
transitions are associated with sequences of symbols from an output
alphabet $\Delta$. It is defined as $T=(\Sigma, \Delta, Q, q_0,
\delta, \sigma, F)$ such that $(\Sigma, Q, q_0,\delta, F)$ is a DFSA
and $\sigma(q, a)$ is the output associated with the transition leaving
$q$ and consuming the input symbol  $a$.

In addition to the concepts introduced above, we will use the
following notation. If $T, T_1, T_2$ are finite-state transducers,
then $T^{-1}$ denotes the result of reversing $T$. $T_1 \cdot T_2$
is the {\em concatenation} of transducers $T_1$ and $T_2$. $T_1 \circ
T_2$ denotes the {\em composition} of $T_1$ and $T_2$.

\section{Requirements}

In this section, we review the state of the art in finite-state
technology from the angle of applicability to the symbolic part of a
TTS system.

\subsection{Finite-State Rule Compilers}

Many solutions have been proposed for compiling rewrite rules into
FSTs, cf. \cite{Kaplan:Kay:94,Roche:Schabes:95,Mohri:Sproat:96}.

Typically, a rewrite rule $\phi \rightarrow \psi / \lambda\_\rho$
states that a string matching a regular expression $\phi$ is rewritten
as $\psi$ if it is preceded by a left context $\lambda$ and followed
by a right context $\rho$, where both $\lambda$ and $\rho$ are stated
as regular expressions over either the input alphabet $\Sigma$ or the
output alphabet $\Delta$.  The compiler compiles the rule by
converting $\phi$, $\lambda$ and $\rho$ into a number of separate
transducers and then composing them into an FST that performs the
rewrite operation.

Since a rule may overlap or conflict with other rules, a
disambiguation strategy is required. There are several
possibilities. Firstly, if the rules are associated with probabilities
or scores, these numeric values may be added to transitions in the
form of weights, thus defining a weighted finite-state transducer
(WFST). Such a WFST is not determinizable in general, but the weights
may be used to guide the search for the best solution and constrain
the search space.

Secondly, a deterministic longest-match strategy may be
pursued. Finally, we may regard the order of the rules as meaningful
in the sense of priorities: if a rule $R_k$ rewrites a string $s$ that
matches its focus $\phi_k$, it blocks the application of all rules
$R_i$ such that $i\geq k$ to any string overlapping with~$s$.

In our research, we have focused on the third strategy as the most
appropriate one in the context of our TTS system and the available
resources. This choice makes determinizability a particularly
desirable feature of the rule FSTs as it guarantees linear-time
processing of input. Although a transducer implementing rules with
unlimited regular expressions in the left and the right context is not
determinizable in general \cite{Poibeau:01}, deterministic processing
is still possible by means of a {\em bimachine},
i.e., an aggregate of a left-to-right and a right-to-left
DFSA \cite{Berstel:79}. For this, the resulting
rule FST must realize a function. 

Unfortunately, the compilers described by \cite{Kaplan:Kay:94} and
\cite{Mohri:Sproat:96} are not guaranteed to produce a functional 
transducer in the general case. Thus, we have had to develop a new,
more appropriate compilation method. The new method is described in
detail in section~\ref{sec:rulecomp}.

\subsection{Complex Input Types}
\label{sec:features}

In rVoice, linguistic information is internally represented by
lists of feature structures. If $o$ is an item and $f$ a feature,
$f(o)$ denotes the value of $f$ on $o$. 

Rewrite operations can be applied to different levels of this model,
the input sequences being either strings of atomic symbols
(characters, phonemes, etc.) or sequences of items characterized by
feature-value pairs. While the former case is straightforward, the
latter requires a translation step from feature structures to a finite
alphabet of symbols.

This issue has been addressed in a wide range of publications. The
solutions proposed mostly guarantee a high degree of expressivity,
including feature unification. The price for the expressive power of
the formalism is non-determinism \cite{Zajac:98} and/or the use of rather
expensive unification operations \cite{Becker:ea:02,Constant:03}.

For efficiency reasons, we have decided to pursue a more modest
approach in the current implementation. The approach is based on the
observation that only a finite number of feature-value pairs are used
in the actual rules. Since distinctions between unseen feature-value
pairs cannot affect the mechanism of rule matching, unseen features
can be ignored and the unseen values of the seen features can be
merged into a special symbol $\#$.

If $f_1\ldots f_K$ are the seen features and $\Sigma_1\ldots \Sigma_K$
the respective sets of values appearing in the rules, then a complex
input item $o$ can be represented by the $K$-tuple $(v_1\ldots v_K)$
such that $v_i \in \Sigma_i \cup \{\#\}$ is defined as

\begin{eqnarray*}
 v_i = \left\{ \begin{array}{l@{\quad:\quad}l} 
	f_i(o) & f_i(o)\in \Sigma_i   \\
	\# & f_i(o)\mbox{ undefined or }f_i(o)\not\in \Sigma_i
\end{array} \right. 
\end{eqnarray*}
The context rules are formulated as regular expressions whose leaves
are {\em item descriptions}.  An {\em item description}, e.g.,
$[pos=nn|nnp\ case=u]$, consists of a set of {\em feature-value
descriptions} (here: $pos=nn|nnp$ and $case=u$), determining a set
$U_j$ of values for the respective feature $f_j$. If no feature-value
description is specified for a feature $f_j$, we set
$U_j=\Sigma_j\cup\{\#\}$. Clearly, an item $(v_1\ldots v_K)$ matches
an item description $[U_1\ldots U_K]$ iff $v_1\in U_1 \ldots v_K\in
U_K$.

This leads to the desired regular interpretation of feature-structure
matching rules: a concatenation of unions (disjunctions) of atomic
values. If {\em case, pos} and {\em type} are the relevant features,
the last one taking values from the set $\{alpha, digit\}$, the item
description $[pos=nn|nnp\ case=u]$ is interpreted as $(nn|nnp) \cdot u
\cdot (alpha|digit|\#)$. Clearly, this interpretation extends to
regular expressions defined over the set of item descriptions. For
example, $([pos=nn|nnp\ case=u])^+$ is interpreted as $((nn|nnp) \cdot
u \cdot (alpha|digit|\#))^+$.

\section{Formalisation} 
\label{sec:rulecomp}

\subsection{The Rule Formalism}

For reasons of readability, we decided to replace the traditional
rule format ($\phi \rightarrow \psi / \lambda \_ \rho$) by
the equivalent notation $\lambda / \phi / \rho \rightarrow \psi$,
which we found much easier to read if $\lambda$ and $\rho$ are complex
feature structures. Thus, the compiler expects an ordered set of rules
in the following format.
\[\lambda_i / \phi_i / \rho_i  \rightarrow  \psi_i, i = 1\ldots n\]
$\lambda_i$ and $\rho_i$ are unrestricted regular expressions over the
input alphabet $\Sigma$. The focus $\phi_i$ is a fixed-length
expression over $\Sigma$. The right-hand side of the rule, $\psi_i$,
is a (possibly empty) sequence of symbols from the output alphabet
$\Delta$.

Compared to \cite{Kaplan:Kay:94} and \cite{Mohri:Sproat:96}, the
expressive power of the formalism is subject to two restrictions.
Firstly, the length of the focus ($\phi$) is fixed for each rule,
which is a reasonable assumption in most of the mappings being
modelled. Secondly, only input symbols are admitted in the context of
a rule, which appears to be a more severe restriction than the first
one, but does not complicate the formal description of the considered
phenomena too much in practice.

\subsection{Auxiliary Operations}

In this section, we define auxiliary operations for creating a rule
FST.
\paragraph{\sf accept$\_$ignoring($\beta$,$M$)}

This operation extends an acceptor for a pattern $\beta$ with loops
ignoring symbols in a set $M$ of markers, $M \cap \Sigma =
\emptyset$. In other words, {\sf accept$\_$ignoring($\beta$, $M$)}
accepts $w \in (\Sigma
\cup M)^*$ iff $w$ can be created from a word $u \in \Sigma^*$
that matches $\beta$ by inserting some symbols from $M$ into $u$.

The construction of {\sf accept$\_$ignoring($\beta$,$M$)} is
straightforward: after creating a deterministic acceptor
$A=(\Sigma,Q,q_0,\delta,F)$ for $\beta$, we add the loop $\delta(q,
\mu) = q$ for each $q\in Q$ and $\mu\in M$.

\paragraph{\sf accept$\_$ignoring$\_$nonfin($\beta$,$M$)}

is like {\sf accept$\_$ignoring($\beta$,$M$)} except
that it does not accept symbols from $M$ at the
end of the input string.  For example, {\sf
accept$\_$ignoring$\_$nonfin($a^*$,$\{\#\}$)} accepts $aaaa$
and $\#\#a\#aa$, but not $aaa\#\#\#$.

The construction of this FSA is similar to that of {\sf
accept$\_$ignoring($\beta$,$M$)}. First, we create a deterministic
acceptor $A=(\Sigma,Q,q_0,\delta,F)$ for $\beta$.  Then a
loop $\delta(q, \mu) = q$ is added to $A$ for each $\mu \in M, q
\not \in F$. Finally for each $q \in F$:
\begin{itemize}
\item if $\delta(q, a)$ is defined, its target is 
	replaced with a new non-final state $q'$;

\item we add the transitions $\delta(q', \mu) := q'$ for each $\mu \in M$
and $\delta(q, \epsilon) := q'$.
\end{itemize}
\begin{figure*}[h]
\begin{center}
\includegraphics[width=4.5cm]{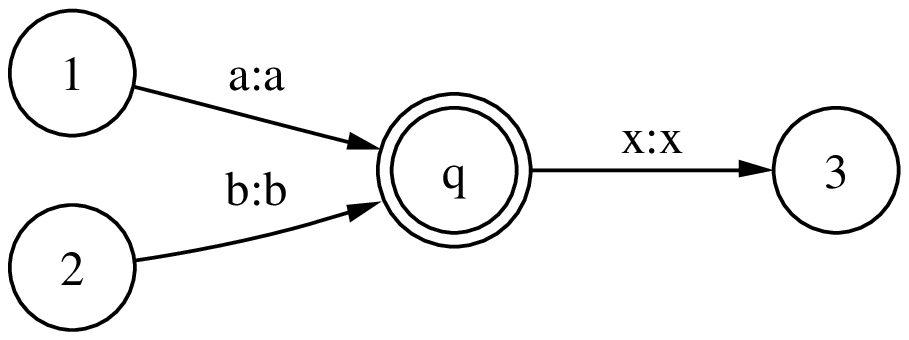}{\raisebox{0.6cm}{\Huge$\Rightarrow$}}
\includegraphics[width=6cm]{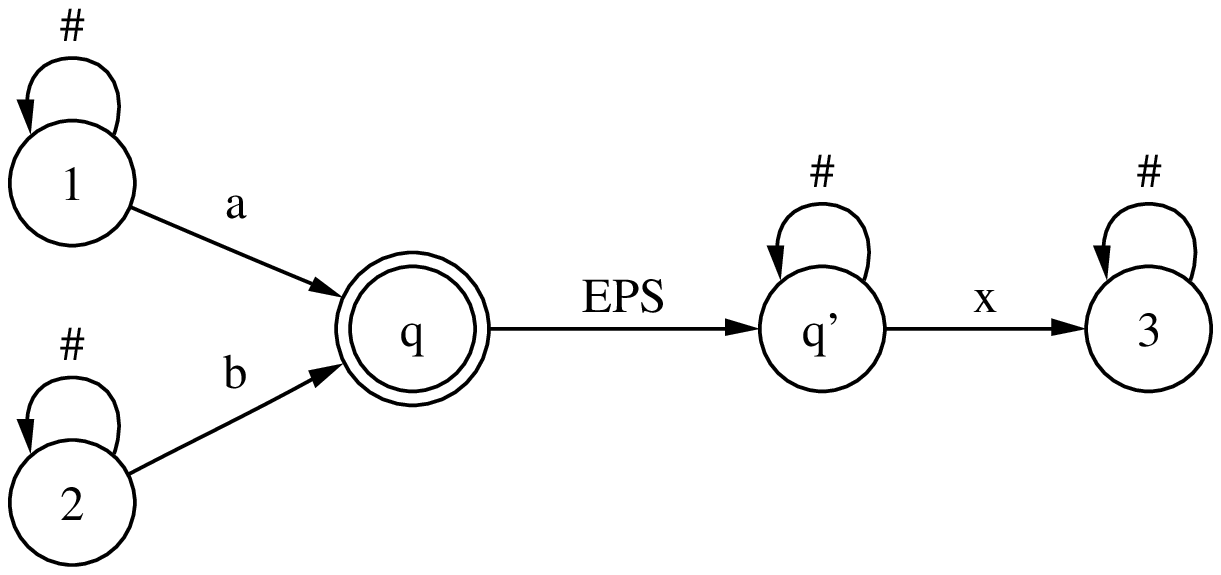}
\end{center}
\caption{\small Construction of {\sf accept$\_$ignoring$\_$nonfin($\beta$,$\{\#\}$)}.}
\label{fig:ignorenonfinal}
\end{figure*}
\paragraph{\sf replace($\beta$,$\gamma$) } 
translates a regular expression $\beta$ into a string
$\gamma$. It is constructed by turning an acceptor
$A=(\Sigma,Q,q_0,\delta,F)$ for $\beta$ into a transducer $T=(\Sigma,Q
\cup \{q_f\}, q_0, \bar{\delta}, \sigma, \{q_f\})$ such that $q_f$ is
a new final state, $\sigma(q,a) := \epsilon$ for each $(q,a) \in
Dom(\delta)$, $\delta \subset \bar{\delta} $,
$\bar{\delta}(q,\epsilon):= q_f$ and $\sigma(q,\epsilon) := \gamma$
for each $q \in F$.

\paragraph{\sf mark$\_$regex$_\Sigma$($\beta$,$\mu$)}

This operation inserts a symbol $\mu$ after each occurrence of a
pattern $\beta$. It is identical to the {\em type 1 marker} transducer 
defined in \cite{Mohri:Sproat:96}. It can be constructed from a deterministic
acceptor $A=(\Sigma,Q,q_0,\delta,F)$ for the pattern $\Sigma^* \beta$
in the following way: first, an identity transducer $Id(A)=
(\Sigma,\Sigma,Q,q_0,\delta,\sigma,F)$ is created such that
$\sigma(q,a) = a$ whenever $\delta(q,a)$ is defined. By construction,
$Id(A)$ is deterministic.

Then, $T=(\Sigma,\Sigma\cup\{\mu\},Q \cup F', q_0,\bar{\delta},\bar{\sigma}, (Q \cup F')\backslash F)$ is created such that

\begin{itemize}
\item[] $F' := \{q' : q \in F\}$ (a copy of each final state of $Id(A)$)
\item[] $\bar{\delta}(q, a) = \delta(q,a), \bar{\sigma}(q, a) = \sigma(q,a)$ for $q \not\in F, a \in \Sigma$
\item[] $\bar{\delta}(q', a) =  \delta(q,a), \bar{\sigma}(q', a) = \sigma(q,a)$ for $q \in F, a \in \Sigma$
\item[] $\bar{\delta}(q, \epsilon) =  q', \bar{\sigma}(q, \epsilon) = \mu$ for $q \in F$
\end{itemize}

\begin{figure*}[h]
\begin{center}
\includegraphics[width=4cm]{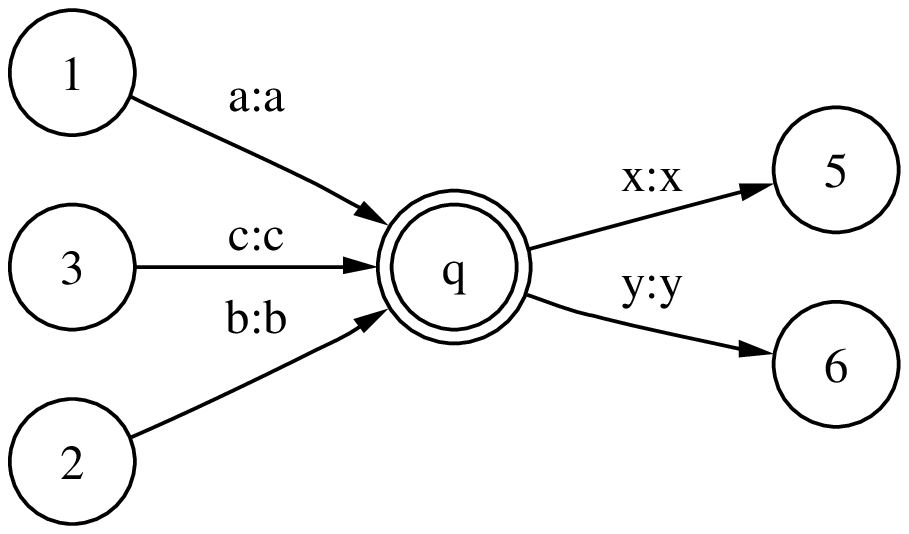}{\raisebox{0.9cm}{\Huge$\Rightarrow$}}
\includegraphics[width=6cm]{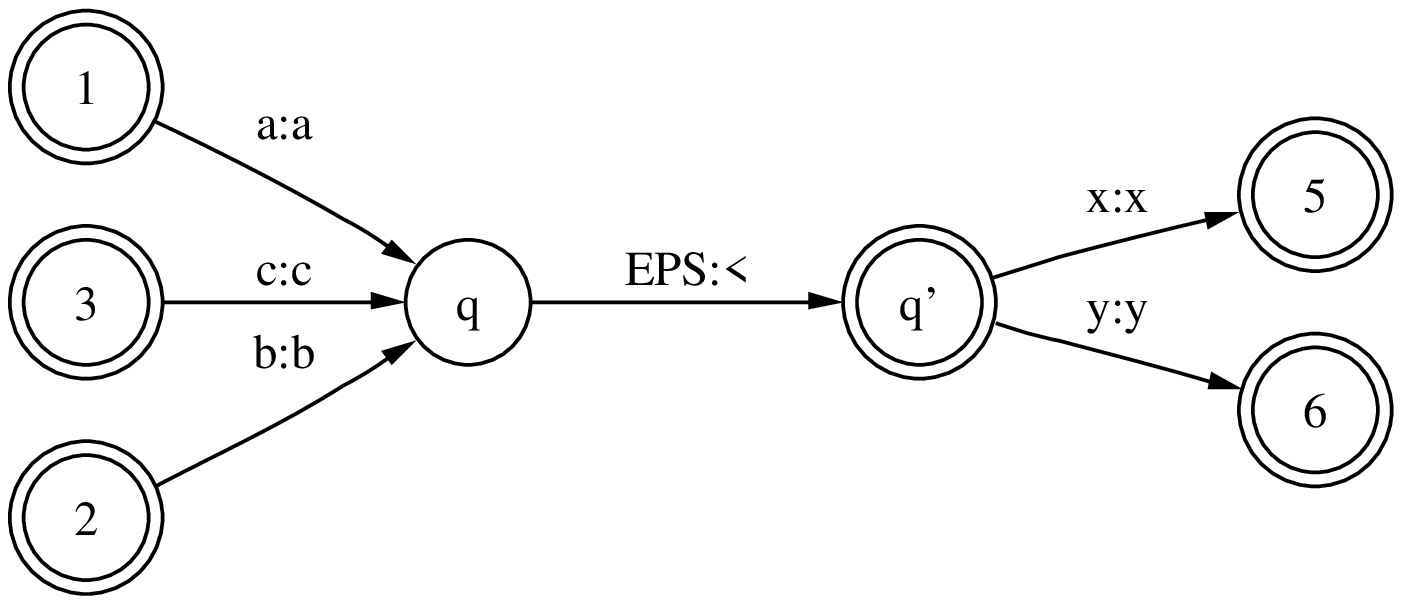}
\end{center}
\caption{\small Construction of the {\sf mark$\_$regex} FST inserting 
$<$ after each match of $\beta$.}
\label{fig:markregex}
\end{figure*}

Informally, the construction of $T$ consists in swapping the final and
non-final states of $Id(A)$ and splitting each final state $q$ of $A$
in two states $q$ and $q'$ such that all transitions $t$ leaving $q$
in $A$ leave $q'$ in $T$. The two states are then connected by a
transition $(q,q', \epsilon, \mu)$, as shown in  figure~\ref{fig:markregex}.

\paragraph{\sf left$\_$context$\_$filter$_\Sigma$($\beta$,$\mu$)}

This operation deletes all occurrences of a symbol $\mu$ in a string
$s \in (\Sigma \cup \{\mu\})^*$ that are {\bf not} preceded by an
instance of pattern $\beta$. A transducer performing this operation
can be constructed from a deterministic acceptor $A=(\Sigma,Q,
q_0,\delta,F)$ for the pattern $\Sigma^* \beta$ by creating an
identity transducer $Id(A) = (\Sigma,\Sigma,Q,q_0,\delta,\sigma,F)$
and then turning it into a transducer $T= (\Sigma \cup \{\mu\},\Sigma
\cup \{\mu\}, Q,q_0,\bar{\delta},\bar{\sigma},Q) $  such
that:

\begin{itemize}
\item[] $\bar{\delta}(q, a) = \delta(q,a), \bar{\sigma}(q, a) = \sigma(q,a)$ for $q \in Dom(\delta)$
\item[] $\bar{\delta}(q, \mu) =  q$ for $q \in Q$
\item[] $\bar{\sigma}(q, \mu) = \mu$ for $q \in F$ (copying of $\mu$ into the 
			output after a match of $\beta$) 
\item[] $\bar{\sigma}(q, \mu) = \epsilon$ for $q \not \in F$(deletion of $\mu$ after a string that does not match $\beta$)
\end{itemize}

\begin{figure*}[h]
\begin{center}
\includegraphics[width=4.5cm]{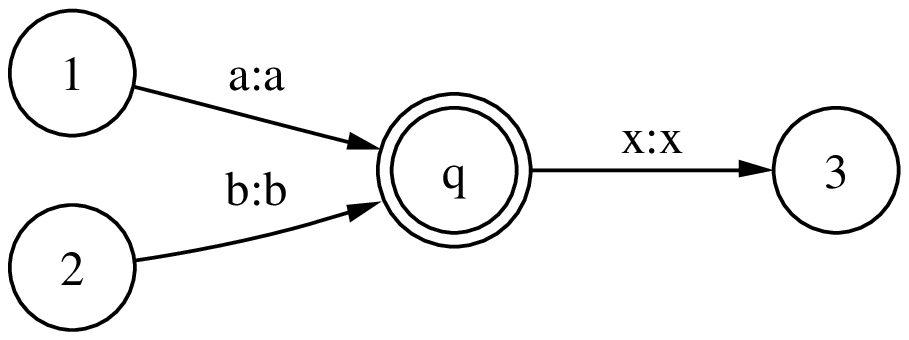}{\raisebox{0.7cm}{\Huge$\Rightarrow$}}
\includegraphics[width=4.5cm]{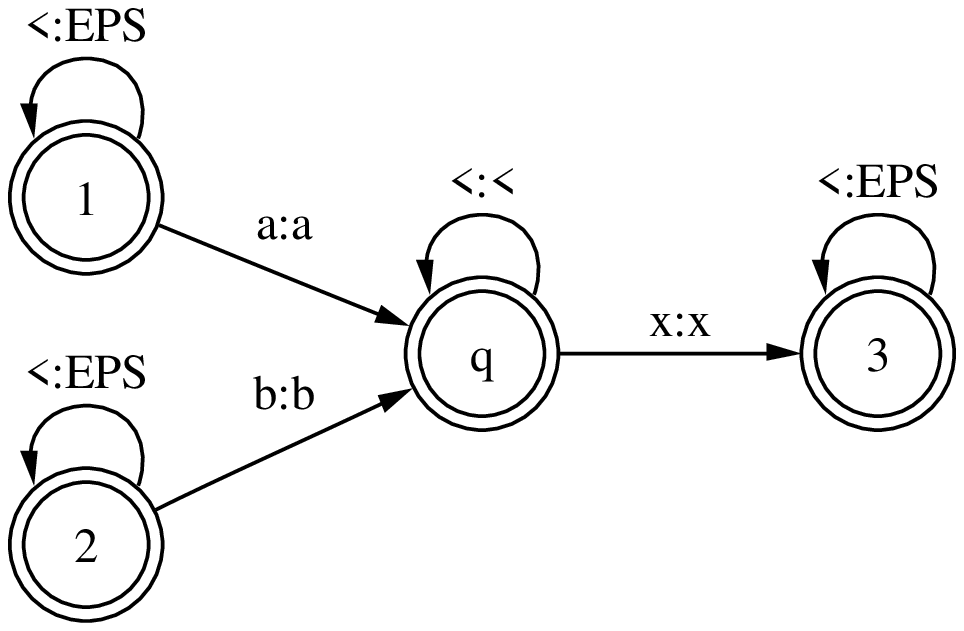}
\end{center}
\caption{\small The {\sf left$\_$context$\_$filter} FST deleting 
$<$ if it is not preceded by $\beta$.}
\label{fig:lcxt}
\end{figure*}

\subsection{Constructing a Rule FST}

Each rule is compiled into a composition of two FSTs. The first one
inserts the symbol $<_i$ before each match of 
$\phi_i \cdot \mathsf{accept\_ignoring}(\rho_i,
\mathsf{Markers}_{<i})$, where $\mathsf{Markers}_{<i}$ is the set of
all markers $<_j$, $j < i$. The second transducer deletes all
occurrences of $<_i$ that are not preceded by an instance of the left
context pattern $\lambda_i$, possibly interspersed with markers
inserted by previous rules ($<_j$). The resulting translation is the
original string with the marker $<_i$ inserted at all positions where
rule $R_i$ fires.

Both FSTs are obviously functional, and so is their composition.

\subsubsection{Marking of the Right Context and Focus Match}

The first transducer {\sf pre\_mark} inserts a left focus marker ($<_i$) {\em before} 
each match of $\phi_i \cdot \mathsf{accept\_ignoring}(\rho_i, \mathsf{Markers}_{<i})$. It
is right-to-left deterministic and can be created by composing the following operations:
\[\mathsf{pre\_mark_i} = 
	\mathsf{mark\_regex}_{\Sigma \cup \mathsf{Markers}_{<i}}
([\phi_i \cdot \mathsf{accept\_ignoring}(\rho_i, \mathsf{Markers}_{<i})]^{-1}, <_i)^{-1}\]

Note that the {\sf mark$\_$regex} operation is performed relative to
the extended alphabet $\Sigma \cup \mathsf{Markers}_{<i}$ as the input string
may already contain markers inserted by an earlier rule.

\subsubsection{Checking the Left Context}

The task of the second FST, {\sf check\_left\_cxt}, is to
remove all occurrences of $<_i$ that are not preceded by an instance
of $\lambda_i$. Note that the substring matching $\lambda_i$ may
contain some of the markers $<_1,..., <_i$, therefore the {\sf
left$\_$context$\_$filter} operation is performed relative to the
extended alphabet $\Sigma \cup \mathsf{Markers}_{<i} = \Sigma \cup
\{<_1, ..., <_{i-1} \}$.

\[\mathsf{check\_left\_cxt_i} = \mathsf{left\_cxt\_filter}_{\Sigma \cup \mathsf{Markers}_{<i}}
	(\mathsf{accept\_ignoring}(\lambda_i, \mathsf{Markers}_{<i}),<_i)\]

\subsubsection{Composition of Rule Transducers}

The transducer for rule $R_i$ is the result of the composition:
$r_i := \mathsf{pre\_mark_i} \circ \mathsf{check\_left\_cxt_i}$.

Since both transducers are deterministic (hence functional), the result
of their composition is functional, too.  The application of the rules
$R_1,..., R_n$ to a string $s$ is then modelled by the composition of
FST's: $(r_1 \circ r_2 \circ \dots \circ r_n \circ
\mathsf{rewrite})(s)$. {\sf rewrite} is a simple FST that, having
read a marker symbol $<_i$, leaves the initial state and jumps to a
subnetwork translating $\phi_i$ to $\psi_i$ (ignoring markers). When 
the translation is finished, {\sf rewrite} returns to its
initial state. {\sf rewrite} can be constructed as the closure of the
union of transducers {\sf rewrite$\_$rule$\_$focus$_i$}, $i=1...n$,
defined as:
\footnote{We assume that at least one marker will be inserted at each position 
in the input string. This can be achieved by specifying
a default rule $/ \mu / \rightarrow \gamma$ for each $\mu \in \Sigma$.}
\[
\mathsf{rewrite\_rule\_focus_i} := \mathsf{replace}(<_i \cdot \mathsf{accept\_ignoring\_nonfin}(\phi_i, \mathsf{Markers}_{\geq i}), \psi_i)
\]
Note the use of {\sf accept$\_$ignoring$\_$nonfin}
rather than just {\sf accept$\_$ignoring}.  This guarantees that the
transducer will not consume any markers following the last character
of $\phi_i$ (these markers indicate the next rule application).

The transducer {\sf rewrite} is then defined as:
\[\mathsf{rewrite} := (\bigcup_{i=1}^n \mathsf{rewrite\_rule\_focus_i}) ^*\]
Clearly, {\sf accept$\_$ignoring$\_$nonfin} is determinizable, and the
resulting transducer {\sf rewrite} is deterministic. With $r_1\ldots
r_n$ being functional, it follows that the rational relation $r_1
\circ r_2 \circ \dots \circ r_n \circ\mathsf{rewrite}$ is functional.
Therefore, the result of the compilation is a functional FST that is
either determinizable or can be factorized into a bimachine.
\footnote{Note that if the focus of a rule contains more than one
character, rules with lesser priority may insert markers into the
matched string. For example, the rules $R_1: / a b / -> X$ and $R_2: /
b / -> Y$ will mark up the string $ab$ as $<_1a<_2b$, but, in
accordance with the operational semantics of the compiler, the
second marker will be ignored by the {\sf rewrite} transducer when the
match is rewritten as $X$.  }

\section{Applications}

rVoice is implemented as a pipeline of modules that successively
transform the input string into sound. The text processing modules
create a sequence of {\em segments} (phones and pauses) organized into
syllables, words and phrases. The result is passed to the {\em speech
modules} that generate the actual speech signal. At each level,
linguistic information is represented by a heterogeneous relation
graph \cite{Taylor:Black:Caley:01}, typcally a list of feature
structures.

Each module creates a new relation or adds information to the existing
ones. The tokenizer splits the input string into a list of tokens. The
text normalisation module expands abbreviations, numbers, dates, etc.,
creating a list of words, each one annotated with a normalised word
form. Further modules (part-of-speech/homograph tagger, reduction
module, language identification) set features such as part-of-speech
on the words.

The lexicon module tries to find a phonetic transcription for the
normalized word form that is consistent with the features set on
it. If it fails, the transcription is generated by letter-to-sound
rules.


In order to acommodate the requirements of different TTS modules, our
rule compiler is parameterizable with respect to input types and
emissions. Two specific instantiations have been employed so far.  The
first one is the conversion the string of atomic symbols from an input
alphabet $\Sigma$ into a string of symbols from an output alphabet
$\Delta$. The second application is setting features on a list of
complex objects ({\em relation items}). In the remainder of this
section, we illustrate each of the two scenarios with an application.

\subsection{Grapheme-To-Phoneme Conversion}

The case of grapheme-to-phoneme conversion is straightforward.  The
input alphabet $\Sigma$ comprises all alphabetic characters, while the
output alphabet $\Delta$ is the set of all legal phonetic symbols of
the language under consideration. For each character, we need to write
rules describing how this character can be pronounced. If more than
one pronunciation is possible, each variant is covered by a rule. The
ordering of the rules determines how conflicts between rules are
resolved and makes it possible to write simple default rules.

The following rules describe the pronunciation of `c' in 
American Spanish:
{\tt\small
\begin{verbatim}
    /  s c / (e|i) -> s ;  # ascienda -> [ a s i e n d a ]
    /  c / (e|i) -> s ;    # cenar    -> [ s e n a r ] 
    / c h / -> ch ;        # ocho     -> [ o ch o ]
    / c / -> k ;           # default rule: k
\end{verbatim}
} Such hand-written rules are used for languages that have a very
regular orthography, such as Spanish, which is covered by 110 rules,
including stress assignment. The resulting FST has 119 states and 5160
transitions.

\subsection{Homograph Disambiguation}

In rVoice, homograph disambiguation is the result of an interaction
between several modules. First of all, a statistical part-of-speech
tagger determines the part-of-speech of each word in an
utterance. This information is useful, but not always sufficient for
determining the right pronunciation. For instance, both pronunciation
variants of {\em lead} are compatible with the POS {\em noun}, as in
the sentences {\em Lakeview took a 14-0 lead in the second quarter}
and {\em There's very high lead levels in your water}. Furthermore,
the POS tagger may be consistently inaccurate in certain contexts, in
which case its predictions are overridden by hand-written homograph
rules. The rules refer directly to the sense IDs associated with the
pronunciation variants of the word in question, as the rules
that disambiguate between the different senses of \emph{suspects}:
{\tt\small
\begin{verbatim}
  [name=that] / [name=suspects] / -> [sense=2];
  ([pos=dt|cd]|[name=terror]) / [name=suspects] / -> [sense=1];
   / [name=suspects] / [name=that] -> [sense=2];
  / [name=suspects] / -> [sense=1]; # default rule
\end{verbatim}
} 
To explain how the rules interact, we will look at the following example:
\begin{center}
\emph{the}$_1$ \emph{terror}$_2$ \emph{suspects}$_3$ \emph{that}$_4$  \emph{were}$_5$  \emph{in}$_6$ \emph{court}$_7$\\
\end{center}
We can see that the second and the third rule  match the context
of word~3. The action associated with the lower rule index is chosen,
resulting in the value of \emph{sense} being set to 1 on the
item. 

According to the compilation method described in
section~\ref{sec:features}, a sequence of items is translated into a
sequence of relevant feature values. The compiled rule FST
rewrites this sequence as a sequence of features to be set according
to the right-hand-side of the rules (in this case, it is the feature
{\em sense}).

\section{Conclusions}

By using FSTs, we have achieved a uniform and declarative way of
expressing linguistic knowledge in rVoice. The rule compilers are run
off-line for each FST-based module, producing a DFST encoding the
combined rules used by this particular module. The FST is loaded by
the system at runtime.  Thus, it has been possible to achieve a clear
separation of the language-independent processing algorithms and the
language-, accent- or speaker-specific data (the FSTs). The (minimized
and determinized) FSTs have contributed to a significant speedup and
footprint reduction.

The interaction of the rule-based FSTs and the automatically trained
text modules (POS tagger, language identification) reflects the
strengths of both approaches. The latter components, trained on
newspaper text, guarantee a high accuracy baseline on input similar to
the available training material. In particular, the POS accuracy is
over 96\% on news text, while the accuracy of language identification
exceeds 99\% (both measured per token). The rule-based modules are
typically used to correct or to complement the predictions of the
automatically trained modules, for example on untypical text genres,
or in response to specific customer requirements.

The FST-based rVoice modules comprise homograph disambiguation,
post-lexical reductions, grapheme-to-phoneme conversion and
syllabification. In all these applications, the compiler has proved to
be a flexible and useful component of the system.

\bibliography{../references/references}
\end{document}